\documentclass{article}
\usepackage{spconf,amsmath,graphicx,hyperref}
\usepackage{amssymb}
\usepackage{booktabs}

\title{Never Too Rigid to Reach: Adaptive Virtual Model Control with LLM- and Lyapunov-Based Reinforcement Learning}
\name{Jingzehua Xu$^{1,\dagger}$ \qquad Yangyang Li$^{2,\dagger}$ \qquad Yangfei Chen$^{3}$ \qquad Guanwen Xie$^{4}$ \qquad Shuai Zhang$^{5}$ \thanks{$^\dagger$ These authors contribute to this work equally.}}

  \address{$^{1}$ Department of Engineering, University of Cambridge \\
      $^{2}$ Department of Electrical Engineering and Computer Science, Massachusetts Institute of Technology \\
      $^{3}$ Zhejiang University-University of Illinois Urbana-Champaign Institute, Zhejiang University \\
      $^{4}$ Tsinghua Shenzhen International Graduate School, Tsinghua University \\
      $^{5}$ Department of Data Science, New Jersey Institute of Technology}
\begin{document}
\ninept
\maketitle
\begin{abstract}
Robotic arms are increasingly deployed in uncertain environments, yet conventional control pipelines often become rigid and brittle when exposed to perturbations or incomplete information. Virtual Model Control (VMC) enables compliant behaviors by embedding virtual forces and mapping them into joint torques, but its reliance on fixed parameters and limited coordination among virtual components constrains adaptability and may undermine stability as task objectives evolve. To address these limitations, we propose \textbf{Adaptive VMC with Large Language Model (LLM)- and Lyapunov-Based Reinforcement Learning} (RL), which preserves the physical interpretability of VMC while supporting stability-guaranteed online adaptation. The LLM provides structured priors and high-level reasoning that enhance coordination among virtual components, improve sample efficiency, and facilitate flexible adjustment to varying task requirements. Complementarily, Lyapunov-based RL enforces theoretical stability constraints, ensuring safe and reliable adaptation under uncertainty. Extensive simulations on a 7-DoF Panda arm demonstrate that our approach effectively balances competing objectives in dynamic tasks, achieving superior performance while highlighting the synergistic benefits of LLM guidance and Lyapunov-constrained adaptation.
\end{abstract}
\section{Introduction}
Robotic arms are a cornerstone of modern automation, supporting applications from precision manufacturing and medical interventions to domestic assistance and human–robot collaboration \cite{1,3}. These applications highlight their societal importance, as manipulators extend human capabilities by enabling accurate, efficient, and safe physical interaction \cite{4}. However, when deployed beyond structured industrial settings into dynamic and uncertain environments, control becomes far more challenging \cite{5}. In such contexts, robots must not only ensure precise reaching but also maintain compliance under perturbations, modeling inaccuracies, and unforeseen contacts \cite{6}. Conventional pipelines that integrate perception, planning, inverse kinematics, and high-gain control, though effective in structured scenarios, often become rigid and unreliable under partial information or unexpected disturbances \cite{7}. This rigidity limits adaptability and may even cause unsafe behaviors, underscoring the need for control frameworks that jointly balance accuracy, robustness, and compliance \cite{31}.

To address this challenge, researchers have explored frameworks that embed physical priors into robotic systems \cite{8}. Among them, Virtual Model Control (VMC) has attracted attention by introducing virtual components—such as springs and dampers—into task space and mapping the resulting forces into joint torques via Jacobian transformations \cite{9}. This formulation naturally couples planning and control while preserving passivity and impedance-shaping properties, thus yielding intuitive and compliant behaviors well suited to uncertain environments \cite{10}. Yet, conventional VMC depends on fixed parameters and exhibits limited coordination among virtual components, which constrains adaptability and may undermine stability as task objectives or environmental conditions evolve \cite{11}.

Fortunately, Reinforcement Learning (RL) has emerged as a principled means of acquiring adaptive policies through interaction with the environment, allowing online adjustment of controller parameters and improving robustness \cite{12}. Nevertheless, end-to-end torque policies often neglect physical priors, resulting in instability, poor sample efficiency, and limited generalization \cite{32}. These drawbacks motivate the integration of RL with structure-preserving frameworks such as VMC, where physical interpretability can be retained while adaptability is enhanced. Building on this idea, recent advances suggest that Large Language Models (LLMs) can provide structured priors and high-level reasoning to guide policy learning, thereby improving coordination among virtual components and supporting flexible task adaptation \cite{25,33}. Meanwhile, Lyapunov-based RL introduces theoretical stability guarantees, ensuring safe and reliable controller adjustment under uncertainty \cite{26, 34}.

Bringing these elements together, we propose \textbf{Adaptive VMC with LLM- and Lyapunov-Based RL}, which combines the physical interpretability of VMC with semantic guidance from the LLM and stability constraints from Lyapunov-based RL. Unlike conventional VMC with fixed parameters and limited component coordination \cite{23}, our framework supports online adaptation of proportional–derivative gains and task-space signals while maintaining physical consistency. Virtual forces generated in this process are mapped through Jacobians and combined with gravity compensation to yield compliant joint torques, enabling robust reaching performance in dynamic and uncertain environments.

\setcounter{figure}{0}
\begin{figure*}[!t]
    \centering
    \includegraphics[width=0.788\linewidth]{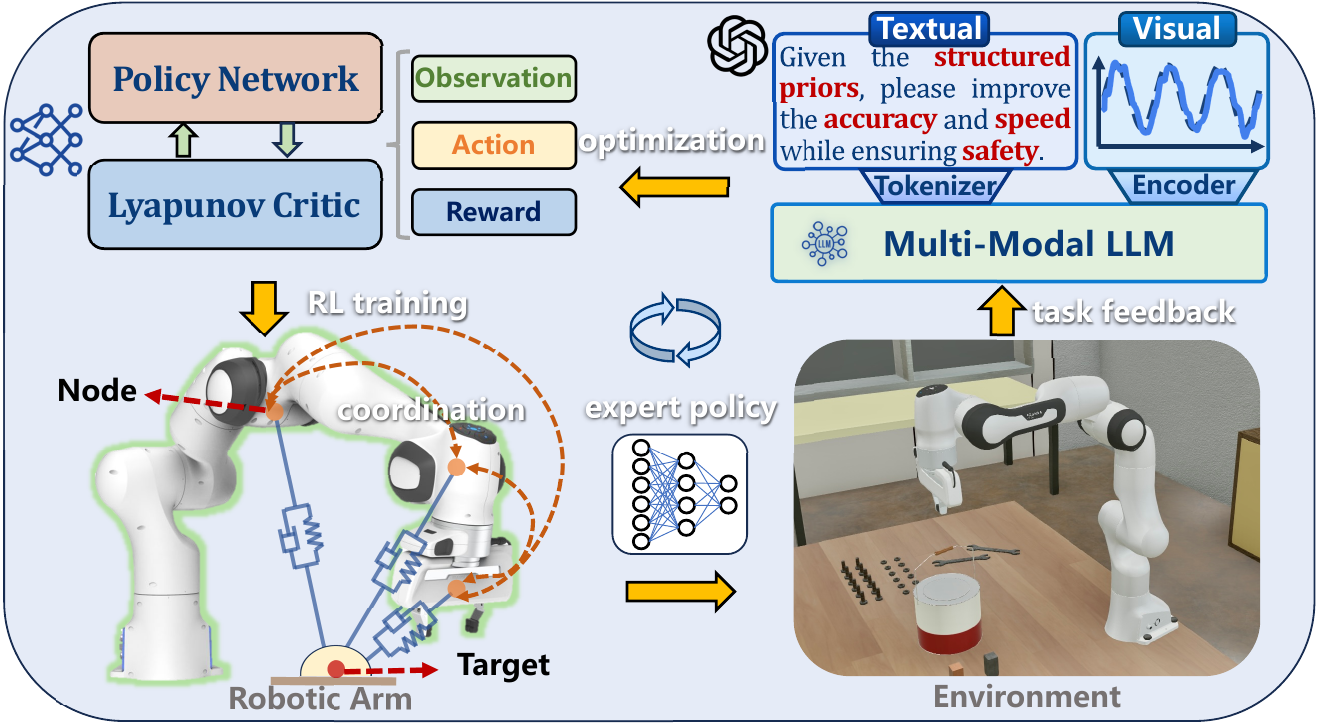}
    \vspace{-3mm}
    \caption{\small Overall architecture of our proposed adaptive VMC framework, which integrates VMC with LLM- and Lyapunov-based RL.} %
    \label{fig_2}
\end{figure*}\vspace{-0mm}

The contributions of this paper are summarized as follows:
\begin{itemize}
\item \textbf{Adaptive VMC Framework}: We propose a novel framework that integrates VMC with LLM- and Lyapunov-based RL, uniting physical interpretability with stability-guaranteed adaptability to meet diverse and changing task requirements.
\item \textbf{LLM-Guided and Lyapunov-Constrained Learning}: LLMs introduce structured priors and reasoning to improve coordination among virtual components, while Lyapunov-based RL enforces theoretical stability guarantees for safe adaptation.
\item \textbf{Extensive Evaluation and Analysis}: We conduct extensive simulations on a 7-DoF Panda arm in Webots, demonstrating that our framework effectively balances competing objectives and achieves superior performance, with results further underscoring the complementary roles of LLM guidance and Lyapunov-constrained adaptation.
\end{itemize}

\section{Methodology}
In this section, we introduce the proposed adaptive VMC framework (as shown in Fig. ~\ref{fig_2}) in detail, which augments VMC with RL for online parameter adaptation and improved component coordination, guided by LLM priors and constrained by Lyapunov stability, enabling adaptive responses to diverse task demands while ensuring compliant and robust reaching under uncertainty.

\subsection{Virtual Model Control with Adaptive Extensions}
VMC generates compliant behaviors by embedding virtual components such as springs and dampers in task space and mapping their forces into joint torques through Jacobian transformations \cite{9}. To formalize this mechanism at the link level, we define the potential and damping energies for each controlled link $i \in \mathcal{V}$ as follows:
\begin{equation}
E_i(\mathbf{p}_i) = \tfrac{1}{2} K_{p,i}|\mathbf{p}_{\mathrm{tar}} - \mathbf{p}_i|^2, \quad
D_i(\dot{\mathbf{p}}_i) = \tfrac{1}{2} K_{d,i}|\dot{\mathbf{p}}_i|^2 ,
\end{equation}
where $K_{p,i}$ and $K_{d,i}$ denote proportional and derivative gains, $\mathbf{p}_{\mathrm{tar}}$ is the target position, and $\mathbf{p}_i$ represents the Cartesian position of link $i$. Then, the corresponding virtual force can be further obtained as
\begin{equation}
\mathbf{F}_i = -K_{p,i}(\mathbf{p}_{\mathrm{tar}}-\mathbf{p}_i) - K_{d,i}\dot{\mathbf{p}}_i .
\end{equation}

These forces are projected into joint torques via the Jacobian $\mathbf{J}_i$:\vspace{-1mm}
\begin{equation}
\boldsymbol{\tau}_\text{task} = \sum_{i \in \mathcal{V}} w_i \mathbf{J}_i^\top \mathbf{F}_i,
\end{equation}
where $w_i$ are adaptive weights balancing the contributions of different links. For clarity, we assume three controlled links ($i \in \{4,6,E\}$), with the weights normalized such that $w_4=\alpha(1-\beta)$, $w_6=\alpha\beta$, and $w_E=1-\alpha$. Finally, the commanded torque includes gravity compensation and actuator saturation:

\begin{equation}
\boldsymbol{\tau} = \mathrm{clip}\left(\boldsymbol{\tau}_\text{task} + \boldsymbol{\tau}_g,,-\tau_{\max},,\tau_{\max}\right).
\end{equation}

This extension preserves the physical intuition of VMC while allowing both the gains $(K_{p,i},K_{d,i})$, and the component weights ${\alpha}$ and $\beta$, to be adaptively optimized, thereby improving compliance and robustness in reaching tasks.

\subsection{Lyapunov-Based Reinforcement Learning}

To optimize the adaptive VMC controller under uncertainty, we adopt an RL formulation where the policy $\pi_\theta$ outputs component gains and coordination weights \cite{30}. The observation $\mathbf{o}_t$ concatenates the Cartesian position, velocity, and target error of selected links, together with the task target:
\begin{equation}
    \mathbf{o}_t = \big[\,\mathbf{p}_i,\,\dot{\mathbf{p}}_i,\,\mathbf{p}_{\mathrm{tar}}-\mathbf{p}_i \,\big]_{i\in\mathcal{V}} \oplus \mathbf{p}_{\mathrm{tar}} .
\end{equation}
Given this observation, the policy $\pi_\theta$, parameterized by a multilayer perceptron, maps $\mathbf{o}_t$ to the corresponding component gains and coordination weights:\vspace{-1mm}
\begin{equation}
    \Theta_t = \{\,K_{p,i}(t), K_{d,i}(t)\,\}_{i\in\mathcal{V}} \cup \{\alpha(t),\beta(t)\},
\end{equation}
which determine the torque through the VMC dynamics.

To ensure stability, we augment the policy with a Lyapunov critic $L_\phi(\mathbf{o}_t)\geq 0$ that enforces the descent condition \vspace{-1mm}
\begin{equation}
    \Delta L = \mathbb{E}\big[L_\phi(\mathbf{o}_{t+1}) - L_\phi(\mathbf{o}_t)\,|\,\pi_\theta\big] \leq -c\|\mathbf{o}_t-\mathbf{o}^*\|^2 ,
\end{equation}
where $\mathbf{o}^*$ is the desired goal state. This condition regularizes policy updates and encourages safe adaptation by ensuring that the learned controller drives the system toward stability \cite{27}.

Building on the above foundation, the objective of RL training is finally formulated to maximize the expected return:
\begin{equation}
    \max_\theta J(\pi_\theta) = \mathbb{E}_{\pi_\theta}\Bigg[\sum_{t=0}^T \gamma^t R_t\Bigg],
\end{equation}
where $\gamma \in [0,1]$ is the discount factor and $R_t$ the instantaneous reward. The reward balances accuracy, efficiency, and safety:
\begin{equation}
    R_t = -\|\mathbf{p}_E-\mathbf{p}_{\mathrm{tar}}\|^2
          -\lambda_\tau \|\boldsymbol{\tau}\|^2
          -\lambda_F \|\mathbf{F}_{\mathrm{rep}}\|^2 ,
\end{equation}
where $\mathbf{p}_E$ denotes the Cartesian position of the end-effector $E$, and the reward penalizes reaching error, excessive torque, and contact forces. The coefficients $\lambda\tau$ and $\lambda_F$ weight control efficiency and contact reduction to ensure safe operation.

\subsection{Large Language Model Guidance}

While RL enables online adaptation of VMC parameters, it often requires extensive exploration and may struggle to encode task semantics or enforce consistent coordination. To mitigate this, we leverage the LLM as a source of structured priors and semantic guidance, ultimately integrating its influence into the reward design \cite{26}.

Given a task description and system feedback $\mathcal{T}$, an LLM encodes semantic information into an embedding
\begin{equation}
    \mathbf{z}_{\mathrm{LLM}} = \mathrm{Enc}_{\mathrm{LLM}}(\mathcal{T}),
\end{equation}
which can be mapped into interpretable task-level cues, such as safety, compliance, or efficiency. These cues are then used to modulate the reward function:
\begin{equation}
    R_t^{\mathrm{LLM}} \!=\! R_t \!-\! \lambda_{\mathrm{rigid}}\mathbf{1}_{\{\text{rigid}\}} 
                              \!-\! \lambda_{\mathrm{unsafe}}\mathbf{1}_{\{\text{unsafe}\}}
                              \!-\! \lambda_{\mathrm{ineff}}\mathbf{1}_{\{\text{inefficient}\}},
\end{equation}
where the indicators $\mathbf{1}_{\{\cdot\}}$ are semantic labels predicted by the LLM.  

In this way, the LLM does not directly control parameters but reshapes the optimization landscape by refining the reward to penalize behaviors misaligned with high-level semantics \cite{29}. For instance, if rigidity is judged to hinder compliance, the weight $\lambda_{\mathrm{rigid}}$ is increased to discourage such behavior during learning.

\setcounter{figure}{1}
\begin{figure}[!t]
    \centering
    \includegraphics[width=0.99\linewidth]{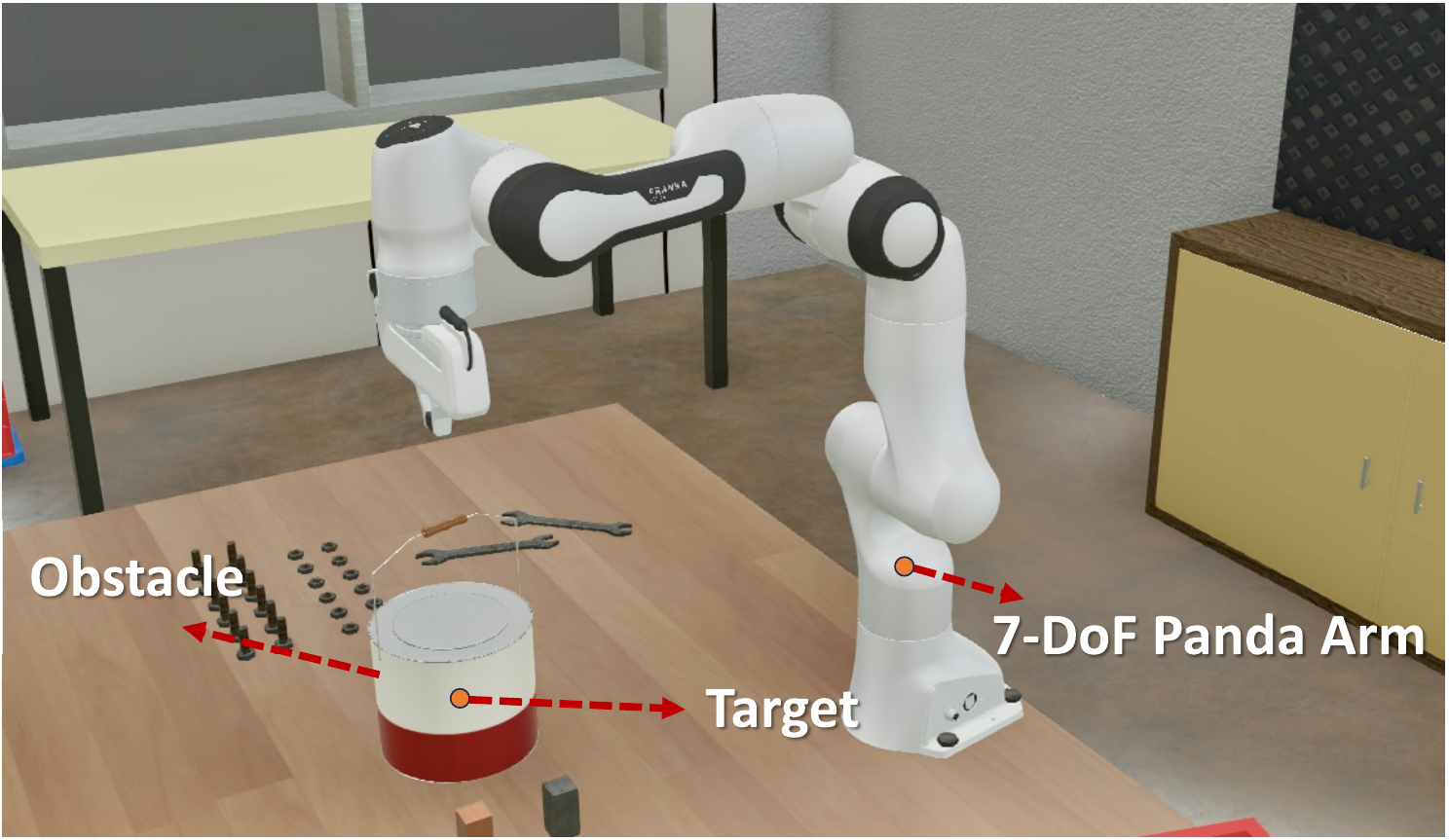}
    \vspace{-2mm}
    \caption{\small Simulation in Webots with a 7-DoF Franka Panda arm.} %
    \label{fig_3_1}\vspace{-2mm}
\end{figure}

Thus, the overall training objective finally becomes
\begin{equation}
    \max_\theta J(\pi_\theta) 
    = \mathbb{E}_{\pi_\theta}\!\left[\sum_{t=0}^T \gamma^t R_t^{\mathrm{LLM}}\right],
\end{equation}
which unifies RL exploration with LLM-informed semantic shaping.  

In summary, LLM guidance is realized through reward design: it injects high-level priors into the learning signal, improving safety, compliance, coordination, and adaptability without requiring manual tuning of control parameters.

\section{Experiments}
In this section, we evaluate the proposed framework through extensive simulations (as shown in Fig. \ref{fig_3_1}), focusing on balancing competing objectives in dynamic tasks and verifying the complementary benefits of LLM guidance and Lyapunov-constrained adaptation.

\subsection{Experimental Setup}
Experiments are carried out in Webots with a 7-DoF Franka Panda arm, where control is distributed over three nodes (links 4, 6, and the end-effector E) \cite{28}. The task requires the end-effector to approach randomized targets while prioritizing safety by avoiding extremely nearby obstacles of varying stiffness. In this setting, each node adaptively adjusts its proportional–derivative gains $(K_p,K_d)$, while two global factors $(\alpha,\beta)$ coordinate their contributions, yielding an 8-dimensional action space. The policy is optimized using PPO with standard hyperparameters \cite{17}, guided by a reward function that penalizes reaching error, torque expenditure, and excessive repulsive forces. Training, averaged over five random seeds, is conducted for 500 episodes of 600 steps each and completes in about 45 minutes on a Ryzen 9 5950X CPU with an RTX 3060 GPU. In summary, our key experimental parameters are summarized in Table~\ref{tab:experiment-parameters}.

\begin{table}[!t]
  \centering
  \caption{\small Key parameters of the experimental setup.}
  \label{tab:experiment-parameters}
  \begin{tabular}{lc}
    \toprule
    \textbf{Parameter} & \textbf{Value \& Description} \\
    \midrule
    Simulation platform & Webots, 7-DoF Panda arm \\
    Controlled nodes & Links 4, 6, end-effector (E) \\
    Action space & $(K_{p,i},K_{d,i})$, $(\alpha,\beta)$ \\
    Policy network & 2-layer MLP (128 units, Tanh) \\
    Lyapunov critic & 2-layer MLP (128 units, Softplus) \\
    PPO settings & lr=$3\times10^{-4}$, $\gamma=0.99$, $\lambda=0.95$ \\
                 & $\epsilon=0.2$, batch=2048, minibatch=256 \\
    LLM model & GPT-4o, temp=0.1, max tokens=300 \\
    Time step & $1/240$ s \\
    Episode and steps & $500$, $600$ \\ 
    Torque limit & $\pm 120$ Nm \\
    Reward Weight & $\lambda_\tau=10^{-4}$, $\lambda_F=10^{-4}$ \\
    \bottomrule
  \end{tabular}
\end{table}

\setcounter{figure}{2}
\begin{figure}[!t]
    \centering
    \includegraphics[width=0.99\linewidth]{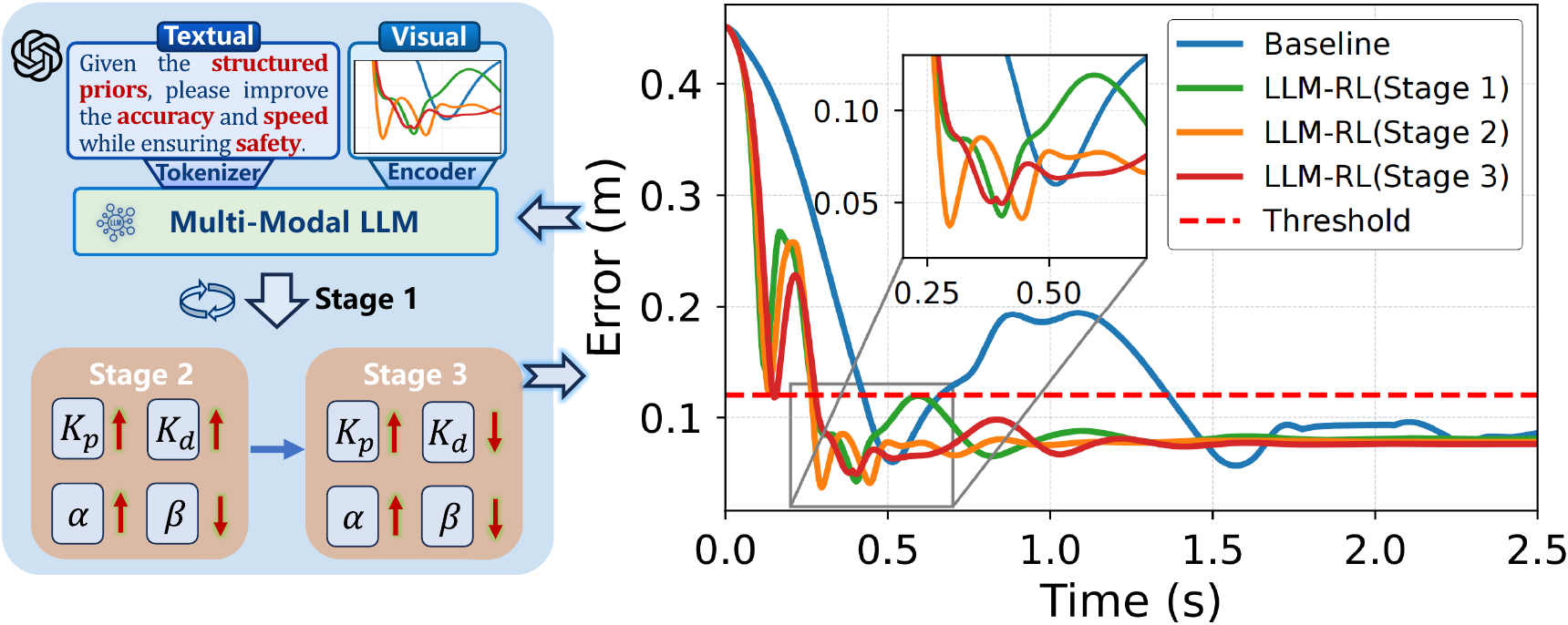}
    \vspace{-3mm}
    \caption{\small Evaluation of our framework compared with the baseline.} %
    \label{fig_3_2}\vspace{-2mm}
\end{figure}

\setcounter{figure}{3}
\begin{figure}[!t]
    \centering
    \includegraphics[width=0.99\linewidth]{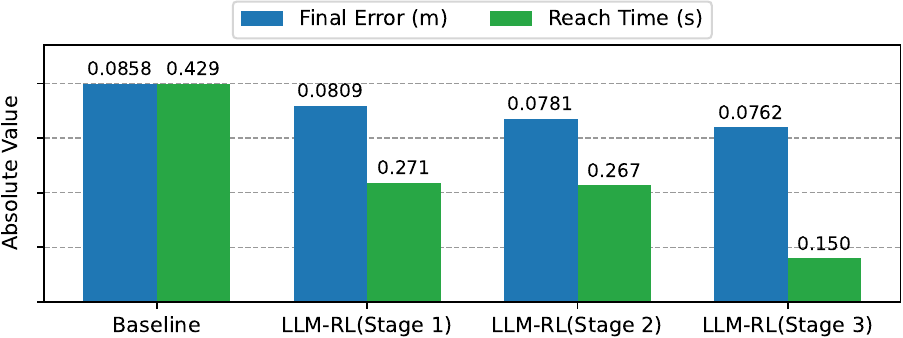}
    \vspace{-3mm}
    \caption{\small Quantitative comparison of our framework with the baseline.} %
    \label{fig_4}\vspace{-3mm}
\end{figure}

\setcounter{figure}{4}
\begin{figure*}[!t]
    \centering
    \includegraphics[width=0.99\linewidth]{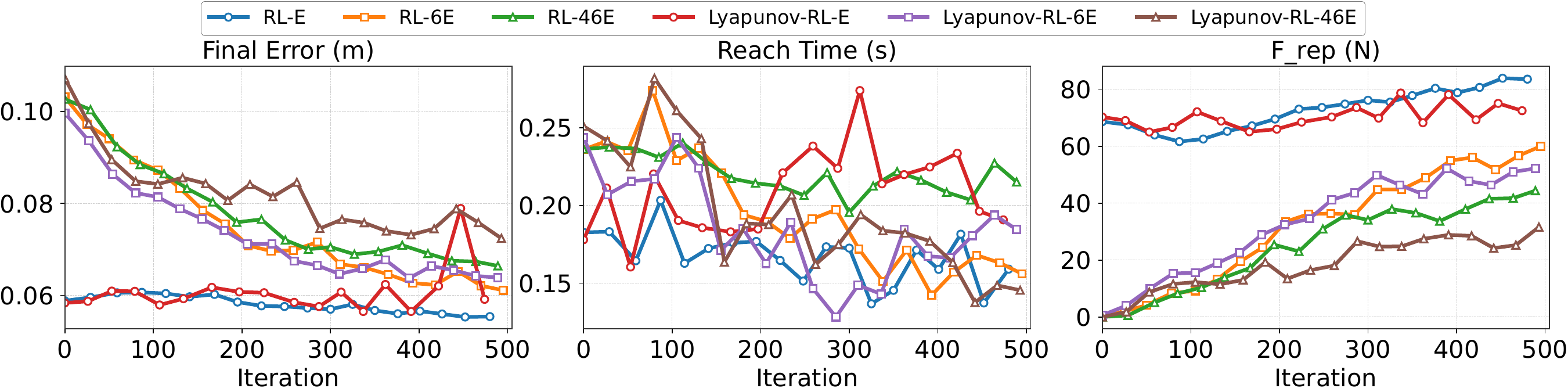}
    \vspace{-2.5mm}
    \caption{\small Learning curves of different VMC configurations under conventional RL and our proposed framework.} %
    \label{fig_5}\vspace{-3mm}
\end{figure*}

\setcounter{figure}{6}
\begin{figure*}[!t]
    \centering
    \includegraphics[width=0.99\linewidth]{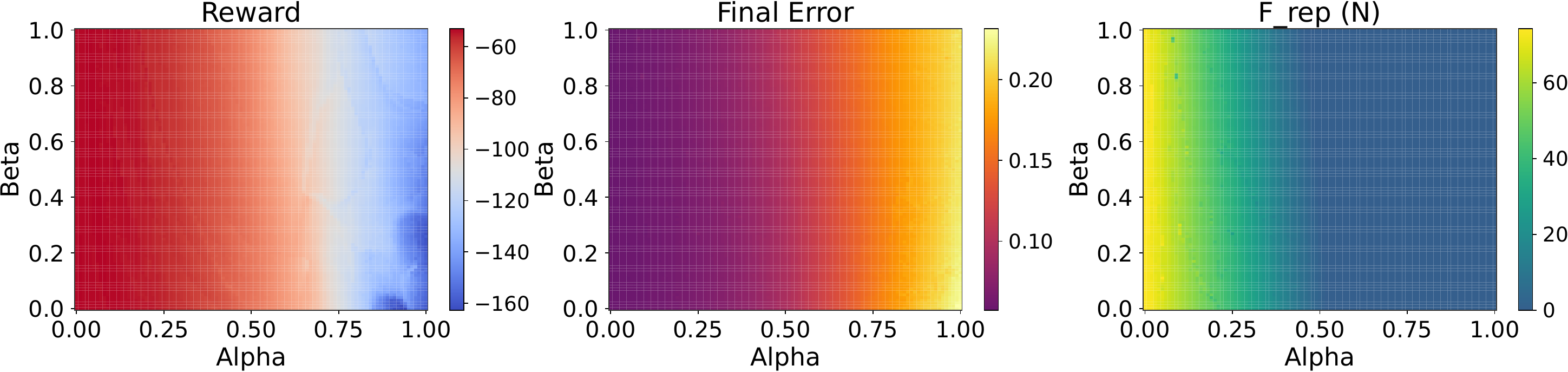}
    \vspace{-2.5mm}
    \caption{\small Heatmaps of coordination factors $(\alpha,\beta)$ under LLM-guided RL, showing their effects on reward, final error, and repulsive force.} %
    \label{fig_6}\vspace{-3mm}
\end{figure*}

\setcounter{figure}{5}
\begin{figure}[!t]
    \centering
    \includegraphics[width=0.99\linewidth]{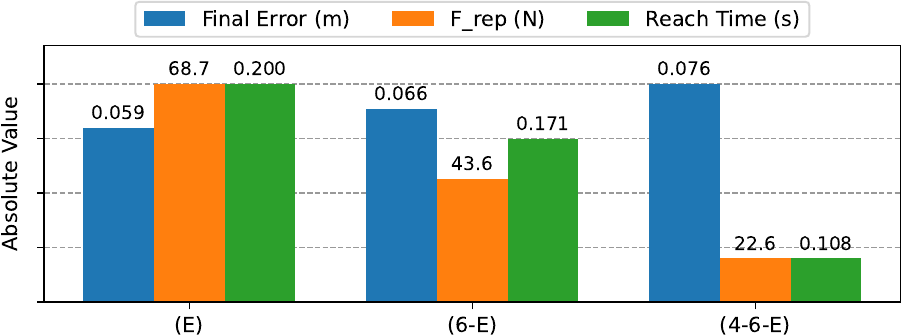}
    \vspace{-2.5mm}
    \caption{\small Comparison of three VMC configurations (E, 6-E, 4-6-E).} %
    \label{fig_7}\vspace{-3mm}
\end{figure}

\subsection{Results and Analysis}
We first evaluate our framework against the baseline method (hand-tuned VMC) from \cite{9}. As shown in Fig. ~\ref{fig_3_2}, the LLM guides parameter adaptation through two successive refinement stages, progressively reducing oscillations and improving compliance, with Stage 3 maintaining errors consistently below the 0.12 m threshold for successful reaching. The quantitative results in Fig. ~\ref{fig_4} further validate these improvements: compared to the baseline (final error 0.0858 m, reach time 0.429 s), our framework achieves a lower error of 0.0762 m and a much shorter reach time of 0.150 s. Overall, the two rounds of LLM guidance enable faster, more stable, and accurate reaching than the baseline, while demonstrating the complementary benefits of semantic priors and Lyapunov-constrained adaptation.

Building upon this baseline comparison, Fig. \ref{fig_5} and Fig. \ref{fig_7} further contrast three VMC configurations: single-node (E), two-node (6-E), and three-node (4-6-E) under both conventional RL and our proposed framework. From the learning curves, single-node RL (E) achieves the lowest final error but suffers from high repulsive forces exceeding 80 N, indicating unsafe trajectories. In contrast, multi-node control reduces repulsive forces significantly while maintaining acceptable accuracy. The bar chart further highlights this trade-off: although 4-6-E has relatively higher error (0.076 m), it achieves the fastest reaching time (0.108 s vs. 0.200 s in E) and the lowest repulsive force (22.6 N). Thus, distributing VMC across multiple nodes slightly sacrifices accuracy but substantially improves safety and efficiency, with 4-6-E configuration offering the best overall balance.

To gain deeper insight into this balance, the heatmaps in Fig.~\ref{fig_6}, obtained under LLM-guided RL training, illustrate the influence of $(\alpha,\beta)$ on reward, error, and repulsive force. The reward landscape shows that low to moderate $\alpha$ values yield higher returns, as stronger end-effector authority enables accurate tracking, whereas performance deteriorates as $\alpha \to 1$ when links 4 and 6 dominate. The error map shows small errors ($<0.1$ m) cluster in the low-$\alpha$ region, while errors above 0.2 m appear as $\alpha$ grows and end-effector contribution weakens. In contrast, the repulsive force map shows safety improves with higher $\alpha$, since reduced end-effector engagement lowers contact forces to near zero, while low $\alpha$ produces forces above 60 N. The role of $\beta$ is minor but provides fine-grained modulation between links 4 and 6. Taken together, these results reveal a tunable trade-off: small $\alpha$ favors accuracy and reward, while large $\alpha$ prioritizes safety, underscoring the role of $(\alpha,\beta)$ in adaptive VMC.

\vspace{-1.5mm}
\section{Conclusions}
In this paper, we propose adaptive VMC with LLM- and Lyapunov-based RL, a framework that balances interpretability and adaptability. The LLM provides structured priors and coordinates virtual components to enhance flexibility and sample efficiency, while Lyapunov-based RL enforces stability through rigorous constraints. Simulations on a 7-DoF Panda arm show that the framework improves accuracy, robustness, and compliance in dynamic tasks. Overall, the results demonstrate the synergistic benefits of combining LLM guidance with Lyapunov-constrained adaptation, offering a reliable path toward safe and adaptive control of robotic manipulators in uncertain environments. Future work will focus on evaluating this framework in more complex tasks and environments, and on releasing the code publicly, aiming to support both academic and industrial development in this area.

\vfill\pagebreak

\bibliographystyle{IEEEbib}
\bibliography{refs}

\end{document}